# A unified setting for inference and decision: An argumentation-based approach

**Leila Amgoud**
IRIT - UPS
118, route de Narbonne
31062, Toulouse, FRANCE
E-mail: amgoud@irit.fr

## Abstract

Inferring from inconsistency and making decisions are two problems which have always been treated separately by researchers in Artificial Intelligence. Consequently, different models have been proposed for each category.

Different argumentation systems [2, 7, 10, 11] have been developed for handling inconsistency in knowledge bases. Recently, other argumentation systems [3, 4, 8] have been defined for making decisions under uncertainty.

The aim of this paper is to present a *general* argumentation framework in which both inferring from inconsistency and decision making are captured.

The proposed framework can be used for decision under uncertainty, multiple criteria decision, rule-based decision and finally case-based decision. Moreover, works on classical decision suppose that the information about environment is coherent, and this no longer required by this general framework.

## 1 INTRODUCTION

Decision making and inference have been studied for a long time separately. Indeed, they have been considered as two distinct problems. Consequently, several models have been proposed for each problem.

The basic idea behind inference is to make conclusions from a set of premises. In other terms, to decide whether a given conclusion is true on the basis of existing information.

The decision problem consists of defining a pre-ordering on a set of possible decisions, on the basis of available information and the goals satisfied or violated by each decision.

A common model for both problems is the one based on argumentation. Argumentation is a reasoning model based on the construction of arguments in favor and against a given statement and then to select the most acceptable of them. Several argumentation systems have been defined for inferring from inconsistent knowledge bases (e.g. [2, 11]). Indeed, conclusions supported by acceptable arguments will be inferred from the knowledge base.

Recently in [3, 4, 8], argumentation has also been used to model the decision making problem. Such an approach has indeed some obvious benefits. On the one hand, it would be more acute with the way humans often deliberate and finally make a choice. Indeed, humans currently use arguments for explaining choices which are already made, or for evaluating potential choices. Each potential choice has usually pros and cons of various strengths. On the other hand, a best choice is not only suggested to the user, but also the reasons of this recommendation can be provided in a format that is easy to grasp. The idea of basing decisions on arguments pro and cons was already advocated more than two hundreds years ago by Benjamin Franklin [9].

In this paper we argue that inference is part of a decision problem. The basic idea is to infer from all the available information, the formulas which are "correctly" supported, then to classify the different decisions on the basis of these formulas.

We propose a general argumentation framework in which the two problems are analyzed and handled. This framework extends the one developed in [1] for inference. The proposed framework is general enough to capture different kinds of decision problems such as decision under uncertainty, multiple criteria decision and rule-based decision. Another feature of the proposed framework is that it extends classical work on decision theory in the sense that the hypothesis that the information about environment is coherent is no longer required by this general framework.

This paper is organized as follows: Section 2 introduces the argumentation process as well as the logical language which will be used throughout the paper. Section 3 presents the different kinds of arguments needed to model inference and decision problems. In section 4 it is shown how arguments can be compared. Section 5 point out the different conflicts which may appear between arguments. In section 6 we define formally an argumentation framework. Section 7 compares our model with existing work on argumentation-based decision making. Section 8 concludes.

## 2 ARGUMENTATION PROCESS

Argumentation is a reasoning model which follows the five following steps:

1. Constructing arguments (in *favor* of / *against* a "statement") from bases.

2. Defining the strengths of those arguments.

3. Determining the different conflicts between the arguments.

4. Evaluating the acceptability of the different arguments.

5. Concluding.

What distinguishes an argumentation framework for reasoning about beliefs and an argumentation framework for decision making is mainly the last step of the argumentation process. Indeed, in inference systems, consequence relations are defined in order to decide which conclusion should be inferred from a knowledge base. Those conclusions are considered "true". However, things seem different with decision making. The basic idea behind a decision problem is to define a pre-ordering, $\triangleright$, on a set $\mathcal{D}$ of possible decisions on the basis of their supporting arguments.

In what follows, let $\mathcal{L}$ be a logical language closed under negation. Rules are given in terms of $\mathcal{L}$ and determine what inferences are possible. A distinction is made between strict rules which will enable to define conclusive inferences and defeasible rules which will enable to define defeasible inferences only. Note that strict rules based on monotonic logic (for example first-order logic) are automatically generated.

- a strict rule is of the form $\phi_1, \ldots, \phi_n \to \phi$. $\mathcal{S}$ will gather all the strict rules.

- a defeasible rule is of the form $\phi_1, \ldots, \phi_n \Rightarrow \phi$. $\mathcal{NS}$ will gather all the defeasible rules.

where $\phi_1, \ldots, \phi_n$ is a finite, possibly empty, sequence in $\mathcal{L}$, and $\phi$ is a member of $\mathcal{L}$.

Let $\mathcal{R} = (\mathcal{S}, \mathcal{NS})$ be the set of rules strict and the defeasible rules. From $\mathcal{L}$ we can distinguish the four following sets:

1. The set $\mathcal{D}$ which gathers all the possible *decisions*.

2. The set $\mathcal{K}$ which represent the *knowledge* base of an agent.

3. The set $\mathcal{G}^+$ which will gather the *positive goals* of an agent. A positive goal represents what an agent wants to achieve.

4. The set $\mathcal{G}^-$ which will gather the *negative goals* of an agent. A negative goal represents what an agent rejects.

**Definition 1 (Theory)** *A theory $\mathcal{T}$ is a tuple ($\mathcal{D}$, $\mathcal{K}$, $\mathcal{G}^+$, $\mathcal{G}^-$).*

We suppose that $\mathcal{K}$ may be pervaded with uncertainty (the beliefs are more or less certain), and the goals in $\mathcal{G}^+$ and $\mathcal{G}^-$ may not have equal priority. Thus, each base is supposed to be equipped with a total preordering $\geq$.

$a \geq b$ iff $a$ is at least as certain (resp. as preferred) as $b$.

For encoding it, we use the set of integers $T = \{0, 1, \ldots, n\}$ as a linearly ordered scale, where $n$ stands for the highest level of certainty or importance and '0' corresponds to the complete lack of certainty or importance. This means that the base $\mathcal{K}$ is partitioned and stratified into $\mathcal{K}_1, \ldots, \mathcal{K}_n$ ($\mathcal{K} = \mathcal{K}_1 \cup \ldots \cup \mathcal{K}_n$) such that all beliefs in $\mathcal{K}_i$ have the same certainty level and are more certain than beliefs in $\mathcal{K}_j$ where j < i. Moreover, $\mathcal{K}_0$ is not considered since it gathers formulas which are completely uncertain, and which are not at all beliefs of the agent.
Similarly, $\mathcal{G}^+ = \mathcal{G}_1^+ \cup \ldots \cup \mathcal{G}_n^+$ and $\mathcal{G}^- = \mathcal{G}_1^- \cup \ldots \cup \mathcal{G}_n^-$ such that goals in $\mathcal{G}_i^+$ (resp. $\mathcal{G}_i^-$) have the same priority and are more important than goals in $\mathcal{G}_j^+$ (resp. $\mathcal{G}_j^-$) where j < i.

**Definition 2 (Closure of a set of formulas)** *Let $F$ be a set of formulas in $\mathcal{L}$. We define the closure of $F$, $Cl_S(F)$, under the set of strict rules $S$ as the smallest set satisfying:*

- $F \subseteq Cl_S(F)$, *and*

- *if* $\phi_1, \ldots, \phi_n \to \psi \in S$ *and* $\phi_1, \ldots, \phi_n \in Cl_S(F)$ *then* $\psi \in Cl_S(F)$.

**Definition 3 (Consistency of a set of formulas)** *Let $F$ be a set of formulas in $\mathcal{L}$. $F$ is consistent iff $Cl_S(F)$ does not contain a formula $\phi$ and its negation $\neg\phi$. Otherwise, $F$ is inconsistent.*

# 3 THE ARGUMENTS

Since decisions are made on the basis of available information on the environment and the goals of the decision maker, two categories of arguments will be defined: *epistemic* arguments for supporting beliefs and *non-epistemic* arguments for supporting decisions. Among non-epistemic arguments, one may distinguish between *recommending* arguments and *decision* arguments. The idea is that a given decision may be justified in two ways: i) it is recommended in a given situation, ii) it satisfies / violates some goals of the decision maker. Recommending arguments have a *deductive* form. Indeed, from formulas, one can deduce using the strict and the defeasible rules a decision. Such arguments are generally constructed for capturing rule-based decision making. Decision arguments have an *abductive* form. These arguments are mainly constructed in case of decision under uncertainty and in multiple criteria decision. In what follows, the set of *argument structures* is defined as follows:

**Definition 4 (Argument structure)** *Let ($\mathcal{D}$, $\mathcal{K}$, $\mathcal{G}^+$, $\mathcal{G}^-$) be different bases.*

- *if $\phi \in \mathcal{K}$ then $\phi$ is an argument structure (A) with:*
  $\text{PROP}(A) = \{\phi\}$
  $\text{GOALS}^+(A) = \emptyset$
  $\text{GOALS}^-(A) = \emptyset$
  $\text{CONC}(A) = \phi$
  $\text{SUB}(A) = \phi$

- *if $A_1, \ldots, A_n$ are argument structures s.t. $\forall A_i$, $\text{GOALS}^+(A_i) = \text{GOALS}^-(A_i) = \emptyset$ and there exits a strict rule $\text{CONC}(A_1), \ldots, \text{CONC}(A_n) \to \psi$ such that $\psi \in \mathcal{K}$ then $A_1, \ldots A_n \to \psi$ is an argument structure (A) with:*
  $\text{PROP}(A) = \text{PROP}(A_1) \cup \ldots \cup \text{PROP}(A_n) \cup \{\psi\}$
  $\text{GOALS}^+(A) = \emptyset$
  $\text{GOALS}^-(A) = \emptyset$
  $\text{CONC}(A) = \psi$
  $\text{SUB}(A) = \text{SUB}(A_1) \cup \ldots \cup \text{SUB}(A_n) \cup \{A\}$

- *if $A_1, \ldots, A_n$ are argument structures s.t. $\forall A_i$, $\text{GOALS}^+(A_i) = \text{GOALS}^-(A_i) = \emptyset$ and there exits a defeasible rule $\text{CONC}(A_1), \ldots, \text{CONC}(A_n) \Rightarrow \psi$ such that $\psi \in \mathcal{K}$, then $A_1, \ldots A_n \Rightarrow \psi$ is an argument structure (A) with:*
  $\text{PROP}(A) = \text{PROP}(A_1) \cup \ldots \cup \text{PROP}(A_n) \cup \{\psi\}$
  $\text{GOALS}^+(A) = \emptyset$
  $\text{GOALS}^-(A) = \emptyset$
  $\text{CONC}(A) = \psi$
  $\text{SUB}(A) = \text{SUB}(A_1) \cup \ldots \cup \text{SUB}(A_n) \cup \{A\}$

- *if $A_1, \ldots, A_n$ are argument structures s.t. $\forall A_i$, $\text{GOALS}^+(A_i) = \text{GOALS}^-(A_i) = \emptyset$ and there exits a strict rule $\text{CONC}(A_1), \ldots, \text{CONC}(A_n) \to d$ such that $d \in \mathcal{D}$ then $A_1, \ldots A_n \to d$ is an argument structure (A) with:*
  $\text{PROP}(A) = \text{PROP}(A_1) \cup \ldots \cup \text{PROP}(A_n) \cup \{d\}$
  $\text{GOALS}^+(A) = \emptyset$
  $\text{GOALS}^-(A) = \emptyset$
  $\text{CONC}(A) = d$
  $\text{SUB}(A) = \text{SUB}(A_1) \cup \ldots \cup \text{SUB}(A_n) \cup \{A\}$

- *if $A_1, \ldots, A_n$ are argument structures s.t. $\forall A_i$, $\text{GOALS}^+(A_i) = \text{GOALS}^-(A_i) = \emptyset$ and there exits a defeasible rule $\text{CONC}(A_1), \ldots, \text{CONC}(A_n) \Rightarrow d$ such that $d \in \mathcal{D}$, then $A_1, \ldots A_n \Rightarrow d$ is an argument structure (A) with:*
  $\text{PROP}(A) = \text{PROP}(A_1) \cup \ldots \cup \text{PROP}(A_n) \cup \{d\}$
  $\text{GOALS}^+(A) = \emptyset$
  $\text{GOALS}^-(A) = \emptyset$
  $\text{CONC}(A) = d$
  $\text{SUB}(A) = \text{SUB}(A_1) \cup \ldots \cup \text{SUB}(A_n) \cup \{A\}$

- *if $A_1, \ldots, A_n$ are argument structures s.t. $\forall A_i$, $\text{GOALS}^+(A_i) = \text{GOALS}^-(A_i) = \emptyset$ and there exits a strict rule $\text{CONC}(A_1), \ldots, \text{CONC}(A_n), d \to \psi$ such that $d \in \mathcal{D}$ and $\psi \in \mathcal{G}^+$ then $A_1, \ldots A_n, d \to \psi$ is an argument structure (A) with:*
  $\text{PROP}(A) = \text{PROP}(A_1) \cup \ldots \cup \text{PROP}(A_n) \cup \{\psi, d\}$
  $\text{GOALS}^+(A) = \{\psi\}$
  $\text{GOALS}^-(A) = \emptyset$
  $\text{CONC}(A) = d$
  $\text{SUB}(A) = \text{SUB}(A_1) \cup \ldots \cup \text{SUB}(A_n) \cup \{A\}$

- *if $A_1, \ldots, A_n$ are argument structures s.t. $\forall A_i$, $\text{GOALS}^+(A_i) = \text{GOALS}^-(A_i) = \emptyset$ and there exits a defeasible rule $\text{CONC}(A_1), \ldots, \text{CONC}(A_n), d \Rightarrow \psi$ such that $d \in \mathcal{D}$ and $\psi \in \mathcal{G}^+$, then $A_1, \ldots A_n, d \Rightarrow \psi$ is an argument structure (A) with:*
  $\text{PROP}(A) = \text{PROP}(A_1) \cup \ldots \cup \text{PROP}(A_n) \cup \{\psi, d\}$
  $\text{GOALS}^+(A) = \{\psi\}$
  $\text{GOALS}^-(A) = \emptyset$
  $\text{CONC}(A) = d$
  $\text{SUB}(A) = \text{SUB}(A_1) \cup \ldots \cup \text{SUB}(A_n) \cup \{A\}$

- *if $A_1, \ldots, A_n$ are argument structures s.t. $\forall A_i$, $\text{GOALS}^+(A_i) = \text{GOALS}^-(A_i) = \emptyset$ and there exits a strict rule $\text{CONC}(A_1), \ldots, \text{CONC}(A_n), d \to \psi$ such that $d \in \mathcal{D}$ and $\psi \in \mathcal{G}^-$ then $A_1, \ldots A_n, d \to \psi$ is an argument structure (A) with:*
  $\text{PROP}(A) = \text{PROP}(A_1) \cup \ldots \cup \text{PROP}(A_n) \cup \{\psi, d\}$
  $\text{GOALS}^+(A) = \emptyset$
  $\text{GOALS}^-(A) = \{\psi\}$
  $\text{CONC}(A) = d$
  $\text{SUB}(A) = \text{SUB}(A_1) \cup \ldots \cup \text{SUB}(A_n) \cup \{A\}$

- *if $A_1, \ldots, A_n$ are argument structures s.t. $\forall A_i$, $\text{GOALS}^+(A_i) = \text{GOALS}^-(A_i) = \emptyset$ and there exits a defeasible rule $\text{CONC}(A_1), \ldots, \text{CONC}(A_n), d \Rightarrow \psi$ such that $d \in \mathcal{D}$ and $\psi \in \mathcal{G}^-$, then $A_1, \ldots A_n, d \Rightarrow \psi$ is an argument structure (A) with:*
  $\text{PROP}(A) = \text{PROP}(A_1) \cup \ldots \cup \text{PROP}(A_n) \cup \{\psi, d\}$

$$\texttt{GOALS}^+(A) = \emptyset$$
$$\texttt{GOALS}^-(A) = \{\psi\}$$
$$\texttt{CONC}(A) = d$$
$$\texttt{SUB}(A) = \texttt{SUB}(A_1) \cup \ldots \cup \texttt{SUB}(A_n) \cup \{A\}$$

*If* $\texttt{GOALS}^+(A) = \emptyset$ *and* $\texttt{GOALS}^-(A) = \emptyset$ *and* $\nexists d \in PROP(A)$ *then* $A$ *is an* epistemic *argument.*

*If* $\exists d \in PROP(A)$ *then* $A$ *is a* recommending *argument, and* $d$ *is a* recommended *decision.*

*If* $\texttt{GOALS}^+(A) \neq \emptyset$ *or* $\texttt{GOALS}^-(A) \neq \emptyset$ *then* $A$ *is a* decision *argument.*

*Let* $\mathcal{A}_e$ *be the set of all epistemic arguments,* $\mathcal{A}_r$ *the set of all recommended arguments, and* $\mathcal{A}_d$ *the set of all decision arguments. Let* $\mathcal{A} = \mathcal{A}_e \cup \mathcal{A}_r \cup \mathcal{A}_d$.

Note that all the sub-arguments of a recommending argument (resp. a decision argument) are epistemic ones. Formally:

**Property 1** *Let* $A \in \mathcal{A}_d$ *(resp.* $\mathcal{A}_r$*).* $\forall A' \in SUB(A)$ *such that* $A \neq A'$, *then* $A'$ *is an epistemic argument.*

In order to avoid any *wishful thinking*, a goal cannot be used in order to justify a formula.

Unlike beliefs, a given decision may have an argument *in favor* of it and also an argument *against* it which are not necessarily conflicting. Intuitively, an argument is in favor of a decision if that decision leads to the satisfaction of a positive goal. The arguments which recommend decisions are also in favor of that decision. An argument is against a decision if the decision leads to the satisfaction of a negative goal. Hence, arguments PRO a decision stress the *positive consequences* of the decision, while arguments CONS are only focusing on the negative ones. Let's define two functions which return respectively for a given decision the arguments which are in favor of it and the arguments against it.

**Definition 5 (Arguments PRO)** *Let* $d \in \mathcal{D}$ *and* $B \subseteq \mathcal{A}$. $Arg_P(d, B) = \{A \in B \mid \texttt{CONC}(A) = d$ *and* $(\texttt{GOALS}^+(A) \neq \emptyset$, *or* $d \in \texttt{PROP}(A))\}$.

**Definition 6 (Arguments CONS)** *Let* $d \in \mathcal{D}$. $Arg_C(d, B) = \{A \in B \mid \texttt{CONC}(A) = d$ *and* $\texttt{GOALS}^-(A) \neq \emptyset\}$.

**Example 1** *Let* $\mathcal{K} = \{a; d; a \Rightarrow b; d \Rightarrow \neg b\}$. *The following arguments can be built:*

$A_1 : [a]$

$A_2 : [d]$

$A_3 : [A_1 \Rightarrow b]$

$A_4 : [A_2 \Rightarrow \neg b]$

## 4 COMPARING ARGUMENTS

In [2, 11], it has been argued that arguments may have forces of various strengths. Different definitions of the force of an argument have been proposed in [2, 11]. Generally, the force of an argument can rely on the information from which it is constructed.

Epistemic arguments involve only one kind of information: the beliefs. Thus, the arguments using more certain beliefs are found stronger than arguments using less certain beliefs. A certainty level is then associated with each argument.

**Definition 7 (Certainty level)** *Let* $A$ *be an argument in* $\mathcal{A}$, *and let* $H = \texttt{PROP}(A) \cap \mathcal{K}$. *The* certainty level *of* $A$, *denoted* $Cert(A) = min\{j \mid 1 \leq j \leq n$ *such that* $H_j \neq \emptyset\}$, *where* $H_j$ *denotes* $H \cap \mathcal{K}_j$.

Unlike epistemic arguments, arguments in favor of and arguments against decisions involve both *goals* and *beliefs*. Thus, the force of such arguments depends not only on the quality of beliefs used in these arguments, but also on the *importance* of the *satisfied* (resp. *violated*) goals. Note that since recommending arguments involve only beliefs, then their force is defined in the same way as for epistemic arguments.

**Definition 8 (Importance degree)** *Let* $A$ *be an argument in* $\mathcal{A}_d$. *The* importance degree *of* $A$, *denoted* $Imp(A) = j$ *such that* $\phi \in \texttt{GOALS}^+(A) \cap \mathcal{G}_j^+$, *or* $\phi \in \texttt{GOALS}^-(A) \cap \mathcal{G}_j^-$.

The force of an argument is defined then by two values: $Cert$ which represents the certainty of the beliefs and $Imp$ which represents the importance of the goals satisfied/violated by a given decision.

**Definition 9 (Force of an argument)** *Let* $A$ *be an argument. If* $A$ *is an epistemic or a recommended argument, then the* force *of* $A$ *is* $Force(A) = Cert(A)$, *otherwise* $Force(A) = (Cert(A), Imp(A))$.

The forces of arguments will play three roles: i) they allow an agent to compare different arguments in order to select the 'best' ones, ii) they are useful for determining the acceptable arguments among the conflicting ones and iii) they are also used for ordering decisions. In what follows $\succeq$ will denote any preference relation between arguments.

**Notation 1** *Let* $A$, $B$ *be two arguments of* $\mathcal{A}$. *If* $\succeq$ *is a pre-ordering, then* $A \succeq B$ *means that* $A$ *is at least as 'good' as* $B$. $\succ$ *and* $\approx$ *will denote respectively the strict ordering and the relation of equivalence associated with the preference between arguments.*

Since an argumentation framework for decision making may have three categories of arguments: epistemic arguments, recommending arguments and decision arguments, one may show how mixed arguments can be compared, and how arguments of the same category may be compared. Epistemic arguments always take precedence over arguments for decisions. The reason is that a decision cannot be well supported if the beliefs on which it is based are not justified.

**Definition 10 (Epistemic vs arguments for decisions)**
*Let $A$ be an epistemic argument, and $B$ a decision argument (resp. a recommended argument). It holds that $A \succ B$.*

In normative systems, for instance, where recommending arguments are built from laws and obligations, it is natural to prefer a recommending argument to a decision argument.

**Definition 11 (Recommending vs decision arguments)**
*Let $A$ be a recommending argument, and $B$ a decision argument. It holds that $A \succ B$.*

However, in some other applications, even if a decision $d$ is recommended in a given situation, an agent may prefer another decision $d'$ which satisfies its goals provided that the beliefs used to justify $d'$ are more certain than the beliefs used to recommend $d$. Formally:

**Definition 12 (Recommending vs decision arguments)**
*Let $A$ be a decision argument and $B$ be a recommending argument. $A \succeq B$, iff $Cert(A) \geq Cert(B)$ and $B \succeq A$, iff $Cert(B) \geq Cert(A)$.*

Concerning two epistemic arguments (resp. two recommending arguments), one would prefer the argument with the highest certainty level.

**Definition 13 (Epistemic (resp. recommended) arguments)**
*Let $A, B$ be two epistemic arguments (resp. two recommending arguments). $A$ is preferred to $B$, denoted $A \succeq B$, iff $Force(A) \geq Force(B)$.*

Now, let's see how two decision arguments can be compared. Intuitively, a decision is 'good' if, according to the most certain beliefs, it satisfies an important goal. A decision is weaker if it involves beliefs with a low certainty, or if it only satisfies a goal with low importance. In other terms, the force of an argument represents to what extent the decision maker is certain that the decision will satisfy its most important goals. This suggests the use of a *conjunctive* combination of the certainty and the priority of the most important satisfied (resp. violated) goal.

**Definition 14 (Decision arguments)** *Let $A, B$ be two decision arguments. $A \succeq B$ iff $min(Cert(A), Imp(A)) > min(Cert(B), Imp(B))$.*

**Example 2** *Assume the following scale $\{0, 1, 2, 3, 4, 5\}$. Let us consider two decision arguments $A$ and $B$ whose forces are respectively (3, 2) and (1, 5). In this case the argument $A$ is preferred to $B$ since min(3, 2) = 2, whereas min(1, 5) = 1.*

However, a simple conjunctive combination is open to discussion, since it gives an equal weight to the importance of the goal and to the certainty of the set of beliefs. Indeed, one may prefer an argument that is certain but has 'small' consequences, than an argument which has a rather small plausibility, but which concerns a very important goal. So the above criteria may be refined. The aim of this section is to give an idea of how arguments may be compared, and not to present an exhaustive list of criteria.

## 5 DIALECTICAL INTERACTIONS BETWEEN ARGUMENTS

Since the information may be inconsistent, the arguments may be conflicting. Indeed, arguments supporting beliefs may be conflicting. It may also be the case that arguments supporting beliefs conflict with arguments supporting decisions. Finally, arguments supporting decisions can conflict with each others. Three different kinds of conflicts may exist between arguments of the same category and also the mixed conflicts.

**Definition 15 (Rebutting)** *Let $A$ and $B$ be arguments in $\mathcal{A}$. $A$ rebut-attacks $B$ iff $\exists \phi$ such that $\phi \in \text{PROP}(A)$ and $\neg \phi \in \text{PROP}(B)$. $A$ rebut-defeats $B$ iff $A$ rebut-attacks $B$ and not $(B \succ A)$.*

Notice that rebut-attacks are symmetric.

**Definition 16 (Assumption-attacking)** *Let $A$ and $B$ be arguments in $\mathcal{A}$. $A$ assumption-attacks $B$ on $A'$ iff $A$ has a subargument $A'$ with $\text{CONC}(A') = \phi$ and $B$ has an assumption $\neg \phi$. $A$ assumption-defeats $B$ iff $A$ assumption-attacks $B$ on $A'$ and it is not the case that the argument $\neg \phi \succ A'$.*

Note that if an argument $A$ assumption-attacks itself then $\{A\}$ is inconsistent.

**Definition 17 (Pollock-undercutting)** *Let $A$ and $B$ be arguments in $\mathcal{A}$. $A$ undercut-attacks and undercut-defeats $B$ iff $B$ has a subargument $B'$ of the form $B''_1, \ldots, B''_n \Rightarrow \psi$ and $A$ has a subargument $A'$ with $\text{CONC}(A') = \neg \lceil \text{CONC}(B''_1), \ldots, \text{CONC}(B''_n) \Rightarrow \psi \rceil$.*

The three above relations are brought together in a unique definition of "defeat".

**Definition 18 (Defeating)** *Let $A$ and $B$ be arguments. We say that $A$ defeats $B$ iff:*

- *$A$ rebut-defeats $B$,*
- *$A$ assumption-defeats $B$, or*

- *A undercut-defeats B.*

Note that since epistemic arguments are always preferred to decision and recommended arguments, then an epistemic argument cannot be defeated w.r.t *defeat* by a decision argument or a recommended argument.

**Property 2** *It cannot be the case that $\exists\ A \in A_e$ and $\exists\ B \in A_d$ (or $B \in A_r$) such that B defeats A.*

## 6 ARGUMENTATION FRAMEWORK

Once all the basic concepts introduced, we are now ready to define an argumentation framework.

**Definition 19 (Argumentation framework)** *Let $\mathcal{T}$ be a theory. An* argumentation framework *(AF) built on $\mathcal{T}$ is a triple $<\mathcal{A}$, defeat, $\succeq>$ s.t:*

- *$\mathcal{A}$ is the set of arguments (see Definition 4),*
- *defeat is the relation given in Definition 18.*
- *$\succeq$ is a preference relation between arguments.*

Among all the conflicting arguments, it is important to know which are the arguments which will be kept for inferring conclusions and for ordering decisions. In [7], different semantics for the notion of acceptability have been proposed. Let's recall them here.

**Definition 20 (Conflict-free, Defence)** *Let $S \subseteq \mathcal{A}$.*

- *A set $S$ is* conflict-free *iff there exist no $A_i$, $A_j$ in $S$ such that $A_i$ defeats $A_j$.*
- *A set $S$ defends an argument $A_i$ iff for each argument $B \in \mathcal{A}$, if $B$ defeats $A_i$ there exists $C$ in $S$ such that $C$ defeats $B$.*

**Definition 21 (Acceptability semantics)** *Let $S$ be a subset of $\mathcal{A}$.*

- *Admissible: $S$ is an admissible set iff $S$ is conflict-free and $S$ defends collectively all its elements.*
- *Preferred: $S$ is a preferred extension iff $S$ is maximal for set inclusion among the admissible sets of $\mathcal{A}$.*
- *Complete: an admissible subset $S$ of $\mathcal{A}$ is a complete extension iff every argument which is defended collectively by $S$ belongs to $S$.*
- *Stable: a subset $S$ of $\mathcal{A}$ is a stable extension iff $S$ is conflict-free and $S$ defeats each argument which does not belong to $S$.*
- *Grounded: $S$ is the grounded extension iff $S$ is conflict-free and $S$ is the least fixed point of the characteristic function F of $<\mathcal{A}, \mathcal{R}>$ (F: $2^{\mathcal{A}} \rightarrow 2^{\mathcal{A}}$ with $F(S) = \{A$ such that $A$ is defended collectively by $S\})$.*

Let $\mathcal{S} = \{E_1, \ldots, E_n\}$ be the set of all possible extensions under a given semantics.

The above extensions may contain epistemic and non-epistemic arguments. Moreover, each argument which is in an extension, have all its sub-arguments in that extension.

**Proposition 1** *Let $AF = <\mathcal{A}$, defeat, $\succeq>$ be an argumentation framework and $E_i \in \mathcal{S}$. $\forall\ A \in E_i$, $\text{SUB}(A) \in E_i$.*

Once the acceptable arguments defined, the decisions may be compared on the basis of the quality of their supporting arguments, and conclusions may be inferred from a knowledge base.

**Definition 22 (Inferring)** *Let $AF = <\mathcal{A}$, defeat, $\succeq>$ be an argumentation framework. $\psi$ is inferred from $\mathcal{K}$, denoted by $\mathcal{K} \mid\sim \psi$, iff $\forall\ E_i \in \mathcal{S}$, $\exists\ A \in E_i \cap \mathcal{A}_e$ such that $\text{CONC}(A) = \psi$.*
$\text{Output}(AF) = \{\psi \mid \mathcal{K} \mid\sim \psi\}$.

An important result is that the set of all conclusions inferred from $\mathcal{K}$ is consistent.

**Proposition 2** *Let $AF = <\mathcal{A}$, defeat, $\succeq>$ be an argumentation framework. The set $\text{Output}(AF)$ is consistent.*

Elements of $\text{Output}(AF)$ are considered as true. Note that decisions are not inferred. The reason is that one cannot say that a given decision is true or false. A decision may have only acceptable arguments which are against it. In such a situation that decision should be discarded. So, the idea in a decision problem, is to construct the arguments in favor and against each decision. Then among all those argument, only the strong (acceptable) ones are kept and the different decisions are compared on the basis of them. Comparing decisions is an important step in a decision process. Below we present an example of *intuitive* principle which is reminiscent of classical principles in decision.

**Definition 23 (Comparing decisions)** *Let $AF = <\mathcal{A}$, defeat, $\succeq>$ be an argumentation framework and $E$ its grounded semantics. Let $d_1$, $d_2 \in \mathcal{D}$. Let $Arg_P(d_1, E) = (P_1, \ldots, P_r)$ and $Arg_P(d_2, E) = (P'_1, \ldots, P'_s)$. Each of these vectors is assumed to be decreasingly ordered w.r.t $\succeq$ (e.g. $P_1 \succeq \ldots \succeq P_r$). Let $v = min(r, s)$.*
*A pre-ordering $\triangleright$ on $\mathcal{D}$ is defined as follows: $d_1 \triangleright d_2$ iff:*

- *$P_1 \succ P'_1$, or*
- *$\exists\ k \leq v$ such that $P_k \succ P'_k$ and $\forall\ j < k$, $P_j \approx P'_j$, or*

- $r > v$ and $\forall\, j \leq v,\ P_j \approx P'_j$.

The above principle takes into account only the arguments pro, and prefers a decision which has at least one acceptable argument pro which is preferred (or stronger) to any acceptable argument pro the other decision. When the strongest arguments in favor of $d_1$ and $d_2$ have equivalent strengths (in the sense of $\approx$), these arguments are ignored. We can show that modeling decision making and inference in the same framework does not affect the result of inference. Before that, let's define when two argumentation frameworks are equivalent.

**Definition 24 (Equivalent frameworks)** *Let $(\mathcal{D}, \mathcal{K}, \mathcal{G}^+, \mathcal{G}^-)$ be a theory.*
*An argumentation framework $AF = <\mathcal{A}$, defeat, $\succeq>$ is equivalent to another argumentation framework $AF' = <\mathcal{A}'$, defeat', $\succeq'>$ iff:*

- Output$(AF)$ = Output$(AF')$, *or*

- *The pre-ordering $\triangleright$ is equivalent to $\triangleright'$, i.e. for any decisions $d, d' \in \mathcal{D}$, if $d \triangleright d'$ then $d \triangleright' d'$.*

**Proposition 3** *The two argumentation frameworks $<\mathcal{A}_e$, defeat, $\succeq>$ and $<\mathcal{A}$, defeat, $\succeq>$ are equivalent.*

The above proposition means that an argumentation framework in which only epistemic arguments are taken into account will return exactly the same inferences as an argumentation framework is which all the different kinds of arguments are considered.

## 7   RELATED WORKS

There has been almost no attempt at formalizing the idea of basing decisions on arguments pro and cons until now if we except some recent works by Fox and Parsons [8], and by Bonet and Geffner [4]. However, these works suffer from some drawbacks: the first one being based on an empirical calculus while the second one, although more formal, does not refer to argumentative inference.

More recently, in [3] an argumentation framework for decision making under uncertainty has been proposed. It is a counterpart, in terms of logical arguments, of the possibilistic qualitative decision setting (which has been axiomatized both in the von Neumann [5] and in the Savage styles [6]). That possibilistic setting distinguishes between *pessimistic* and *optimistic* attitudes toward risk. This gave birth to different types of arguments in favor of and against a possible choice built from a consistent knowledge base and a consistent goals base in [3]. Indeed, arguments pro are used to capture the pessimistic attitude whereas arguments cons are used to capture the optimistic one. In our framework, this corresponds to the particular case where $\mathcal{R}$ = $\emptyset$ which means that there are no conflicts between arguments. Moreover, the sets of epistemic and recommended arguments are empty ($\mathcal{A}_r = \mathcal{A}_e = \emptyset$). Thus, all the arguments in $\mathcal{A}_d$ are acceptable ($\mathcal{S} = \{A_d\}$). The ordering defined in [3] using a pessimistic attitude is exactly the pre-ordering $\triangleright$ given in definition 23 applied to the argumentation framework $<\mathcal{A}_d, \emptyset, \succeq>$. However, for capturing the optimistic attitude, one should use the same argumentation framework but with the following relation for comparing decisions:

**Definition 25 (Comparing decisions)** *Let $AF = <\mathcal{A}$, defeat, $\succeq>$ be an argumentation framework and $E$ its grounded semantics. Let $d_1, d_2 \in \mathcal{D}$. Let $Arg_C(d_1, E) = (C_1, \ldots, C_r)$ and $Arg_C(d_2, E) = (C'_1, \ldots, C'_s)$. Each of these vectors is assumed to be decreasingly ordered w.r.t $\succeq$ (e.g. $C_1 \succeq \ldots \succeq C_r$). Let $v = min(r, s)$. $d_1 \succ d_2$ iff:*

- $C'_1 \succ C_1$, *or*

- $\exists\, k \leq v$ *such that* $C'_k \succ C_k$ *and* $\forall\, j < k,\ C_j \approx C'_j$, *or*

- $v < s$ *and* $\forall\, j \leq v,\ C_j \approx C'_j$.

## 8   CONCLUSION

In this paper we have presented a general formal framework for decision making and inference. This offers for the first time a coherent setting for argumentation-based inference and decision. Unlike inference framework where only epistemic arguments exist, in a decision framework two categories of arguments can be built: epistemic ones and arguments for decisions. This is not surprising since decisions are based on some available knowledge. The basic idea behind a decision problem is to infer from the knowledge base justified conclusions which may support decisions if any. Then, decisions will be compared on the basis of the strengths of the arguments in favor and against them. Moreover, the above approach can be applied not only to decision under uncertainty, but also to multiple criteria decision problems as well as rule-based decision.